\documentclass[letterpaper]{article} 
\usepackage{aaai2026}  
\usepackage{times}  
\usepackage{helvet}  
\usepackage{courier}  
\usepackage[hyphens]{url}  
\usepackage{graphicx} 
\usepackage{natbib}  
\usepackage{caption} 
\usepackage{algorithm}
\usepackage{algorithmic}
\usepackage{subcaption}
\usepackage{amsmath}
\usepackage{newfloat}
\usepackage{listings}
\DeclareCaptionStyle{ruled}{labelfont=normalfont,labelsep=colon,strut=off} 
\floatstyle{ruled}
\newfloat{listing}{tb}{lst}{}
\floatname{listing}{Listing}
\nocopyright
\title{Compositional Generation for Long-Horizon Coupled PDEs}

\author{
  Somayajulu L. N. Dhulipala\textsuperscript{\rm 1}\thanks{Corresponding author: som.dhulipala@inl.gov},
  Deep Ray\textsuperscript{\rm 2},
  Nicholas Forman\textsuperscript{\rm 2}
}
\affiliations{
  \textsuperscript{\rm 1}Idaho National Laboratory, Idaho Falls, ID 83415, USA\\
  \textsuperscript{\rm 2}University of Maryland, College Park, MD 20742, USA\\[2pt]
  som.dhulipala@inl.gov
}

\begin{document}

\maketitle

\begin{abstract}
Simulating coupled PDE systems is computationally intensive, and prior efforts have largely focused on training surrogates on the joint (coupled) data, which requires a large amount of data. In the paper, we study compositional diffusion approaches where diffusion models are only trained on the decoupled PDE data and are composed at inference time to recover the coupled field. Specifically, we investigate whether the compositional strategy can be feasible under long time horizons involving a large number of time steps. In addition, we compare a baseline diffusion model with that trained using the v-parameterization strategy. We also introduce a symmetric compositional scheme for the coupled fields based on the Euler scheme. We evaluate on Reaction-Diffusion and modified Burgers with longer time grids, and benchmark against a Fourier Neural Operator trained on coupled data. Despite seeing only decoupled training data, the compositional diffusion models recover coupled trajectories with low error. v-parameterization can improve accuracy over a baseline diffusion model, while the neural operator surrogate remains strongest given that it is trained on the coupled data. These results show that compositional diffusion is a viable strategy towards efficient, long-horizon modeling of coupled PDEs.
\end{abstract}

\section{Introduction}

Solving coupled PDE systems is important for many science and engineering applications. However, the extreme computational cost of simulating these systems requires the creation of surrogate models for an efficient use of PDEs in practice. Numerous studies in the past have relied on developing surrogates by leveraging the full coupled data from the PDE systems. However, creating accurate surrogates would then require a large volume of training data. In this study, we leverage diffusion models, and particularly, denoising diffusion probabilistic models (DDPM), to create surrogates for coupled PDEs but by only leveraging the decoupled PDE data for the training. Simulating de-coupled PDE data is easier since it doesn't require evolving two or more fields jointly in time.

For image generation, diffusion models have been extensively applied in the past (for e.g., see \cite{Liu2022ComposableDiffusionArXiv}). Also, there has been research demonstrating the use of diffusion models trained on different image classes to create images composed of two or more classes. Typically, such a composition is performed through the product of experts strategy \cite{hinton2002poe}. Recent studies have utilized the product of experts strategy to compose joint PDE fields but by utilizing diffusion models trained only on the decoupled fields \cite{zhang2025m2pde}. However, such studies have only targeted small time horizons involving only a few time steps. We further study the traditional $\varepsilon$-prediction versus v-parameterization training strategies for diffusion models, highlighting that v-parameterization improves learning under low signal-to-noise regimes. 

We evaluate on two canonical systems\textemdash Reaction–Diffusion and modified Burgers\textemdash with orders-of-magnitude more time steps than prior setups that had only a handful of time steps. We introduce a symmetric sampler based on the Euler scheme in contrast to the prior studies that use order-dependent compositional strategies. We compare our results against a baseline Fourier Neural Operator trained on coupled PDE data. Despite training only on decoupled PDE data, compositional diffusion models recover coupled trajectories with low error. v-parameterization consistently improves over the traditional $\varepsilon$-prediction, while FNO remains the strongest as it is trained on the coupled data. Overall, compositional diffusion is a viable strategy towards efficient, long-horizon modeling of coupled PDEs.

\section{Related work}

DDPMs were first introduced by \cite{ho2020denoising} which are composed of a controlled noising process and training a neural network to denoise the noisy field. They show that the training loss can be reduced to the neural network learning incremental noise that is applied at each diffusion time step. \cite{song2020score} introduced score-based diffusion models which introduce a stochastic differential equation view of the diffusion process. They show that learning the incremental noise at each diffusion step is equivalent to learning the score function of the data. \cite{saharia2022imagen} introduces a $v$ parameterization trick where the diffusion model is trained on a combination of noise and the clean field. This is shown to improve the training stability under small signal to noise ratios and has empirically resulted in a better sample quality.

\cite{Liu2022ComposableDiffusionArXiv} introduced a product of experts (PoE) formulation to compose joint distributions from marginals. This formulation enabled the composition of individually trained diffusion models to construct images with richer features than the conditionals. \cite{Chung2023DPS} introduces strategies for applying diffusion models in the inverse setting where the data can be noisy. Borrowing the ideas from the PoE viewpoint, \cite{zhang2025m2pde} introduced compositional diffusion strategies for solving coupled PDE by only training on the conditional uncoupled PDE components. While this strategy showed promise in providing coupled fields without actually training on the coupled field data, the applications were restricted to short time horizons with only a few time steps. Additionally, the accuracy of the composed fields was quite inferior compared to baseline surrogates like neural operators trained on the coupled data. In this work, we explore the PoE strategy to compose diffusion models and solve coupled PDE problems for larger time steps (1-2 orders of magnitude more). Additionally, we utilize the v-parameterization strategy to improve the accuracy of the coupled fields in reference to a baseline DDPM approach. It is noted that the diffusion models were not trained on coupled PDE data; they are only trained on the conditional or un-coupled PDE data.

\section{Methodology}

We adopted the DDPM methodology to train individual diffusion models on a PDE field conditioned on the other fields. A field here refers to the output of a PDE in a coupled PDE system and it can be either a scalar or a vector field. Following the notation of \cite{zhang2025m2pde}, $\pmb{z} = \{z_1,z_2,\dots,z_N\} \sim p(z_1,z_2,\dots,z_N)$ be the coupled fields. $z_{\neq i}$ denotes the remaining fields except for the $i^\text{th}$ one.  Then, using DDPM, we model $p(z_i|z_{\neq i})$ which is the field $i$ solved for by fixing the other fields during the transient PDE simulation. The noising step in a traditional DDPM is \cite{ho2020denoising}:
\begin{equation}
z_i^t|z_{\neq i} = \sqrt{\bar\alpha_t} ~z_i^0|z_{\neq i}+\sqrt{1-\bar\alpha_t}~\varepsilon
\end{equation}
where $\varepsilon\sim\mathcal N(0,I)$ and the noise schedule is defined by $\bar\alpha_t=\prod_{s=1}^t \alpha_s,\;\alpha_s=1-\beta_s$. The denoising step is defined as:
\begin{equation}
z_i^{t-1}|z_{\neq i}
=\frac{1}{\sqrt{\alpha_t}}\left(z_i^{t}|z_{\neq i}-\frac{1-\alpha_t}{\sqrt{1-\bar\alpha_t}}~\varepsilon_\theta(z_i^{t},t,z_{\neq i})\right)
+\sigma_t \tau
\end{equation}
where $\tau \sim\mathcal N(0,I)$, $\sigma_t=\sqrt{\tilde\beta_t}$, and $\tilde\beta_t=\beta_t\,\frac{1-\bar\alpha_{t-1}}{1-\bar\alpha_t}$. In the above equation, $\varepsilon_\theta(z_i^{t},t,z_{\neq i})$ represents a neural network which learns the incremental noise at each time step $t$ given the current state of the required field $z_i^{t}$ and the conditional fields which are fixed $z_{\neq i}$. The fixed conditional fields can also include initial and boundary conditions required to solve the PDE.

We also adopted the v-parameterization technique proposed by \cite{saharia2022imagen}. The original field and the noise at time $t$ can be expressed as:
\begin{equation}
    v \equiv\ \alpha\ \varepsilon - \sigma~z_i^{0}|z_{\neq i}
\end{equation}
where $\alpha=\sqrt{\bar{\alpha}_t}$ and $\sigma=\sqrt{1-\bar{\alpha}_t}$. A neural network is trained to learn to predict $v_\theta(z_i^{t},t,z_{\neq i})$. From this noise can be recovered as $\hat\varepsilon=\sigma ~z_i^{t}|z_{\neq i}+\alpha\,v_\theta(z_i^{t},t,z_{\neq i})$ and the denoised field can be recovered as $\hat z_i^{0}|z_{\neq i} = \alpha~z_i^{t}|z_{\neq i} - \sigma\,v_\theta(z_i^{t}|z_{\neq i},t,c)$. Using these the denoised field can be estimated starting from random noise. Working with the v-parameterization balances the gradient updates across the time steps. This is important given that during the initial noising steps, the signal to noise ratio low. As such, in practice, this can make training more stable compared to $\varepsilon$-based training and prediction \cite{saharia2022imagen}. 

Given the conditional fields $z_i^{t}|z_{\neq i}$, a PoE type of formulation is used to obtain the joint fields (i.e., coupled fields $\pmb{z}$). In PoE, the joint is expressed as:
\begin{equation}
p(\pmb{z}^{t}) \propto \prod_{i=1}^N p(z_i^{t}|z_{\neq i})
\end{equation}
This implies that $\nabla_{z_i^t} \log p(\pmb{z}^{t}) = \nabla_{z_i^t} \log p(z_i^{t}|z_{\neq i})$ \cite{Liu2022ComposableDiffusionArXiv}. Using this, we can iteratively update the conditional to compose the joint using the DDPM prediction $\hat z_i^{t}|z_{\neq i}$ at each diffusion step. An Euler/Picard step for per-field update at iteration $k$ is:
\begin{equation}
\hat z_i^{t,k+1}|z_{\neq i} = z_i^{t,k}|z_{\neq i} + \lambda~(\hat z_i^{0}|z_{\neq i} - z_i^{t,k}|z_{\neq i})
\end{equation}
where $\lambda$ is a parameter controlling the level of update at each step $k$. The update $(\hat z_i^{0}|z_{\neq i} - z_i^{t,k}|z_{\neq i})$ can be viewed as being proportional to $\nabla_{z_i^t} \log p(z_i^{t}|z_{\neq i})$. Therefore, the update is similar to a Euler step in the score space with step size $\lambda$. Typically, 2-3 Euler iterations over the conditional fields are performed within each diffusion step.

For learning the noise or the v-parameter at each diffusion step we use a UNet with three encoder stages with channels $[C,~2C, 4C]$ and three decoder stages with skip connections; $C$ here is set to 48 \cite{ronneberger2015u}. Residual convolutional blocks are used with FiLM condition for the diffusion time step \cite{perez2018film} and a SiLU activation is used. Spatial self-attentions are also used after each convolutional block in the encoder and decoder layers. For considering the diffusion time step through FiLM conditioning, a sinusoidal embedding is used. The input channels are the noisy field at time $t$, the conditioned fields that are fixed, and any initial and boundary conditions. The time index is also input to the UNet. The output has a single channel which is either the $\varepsilon$ or v-parameter depending upon the DDPM formulation. Note that the two DDPM variants are only trained on 10,000 training points of uncoupled PDE data. Then they are composed to retrieve the joint or coupled solution to the full PDE.

For reference, we also trained a Fourier Neural Operator on the coupled PDE data \cite{li2021fourier}. We used a two-dimensional FNO with 16 Fourier modes each along the width and height axis. The input channels are the time and spatial points and also the initial conditions. The output channels are the coupled PDE solutions.

\section{Results}

We present two test cases herein. The first test case is the reaction-diffusion equation with two coupled variables that are dependent on time \cite{zhang2025m2pde}. We discretize the spatial domain into 20 points and the time domain into 500 points, which is large compared to the 10 time points used by \cite{zhang2025m2pde}. The second test case is a modified Burgers equation with reaction with two coupled variables that are time dependent \cite{raissi2019physics}. We discretize the spatial domain into 128 points and the time domain into 100 points. We used the epsilon and v-parameter training strategies to train on the de-coupled field data. For this, we developed two DDPMs, one each for the two coupled variables, for each training strategy. We trained the DDPMs on the de-coupled data using 10,000 training points. Then we used the iterative coupling strategy to compose the joint or coupled fields. We used 500 diffusion steps and 2 coupling iterations within each diffusion step. The de-coupled training data consists of the inputs which are the conditional field and the initial condition. These are randomly generated from a Gaussian random field (GRF). Then, the de-coupled system is solved for the outputs. For the FNO training data, the initial conditions for the two fields are generated through a GRF.

\begin{figure*}[htb!]
    \centering
    \begin{subfigure}[b]{0.49\textwidth} \centering
        \includegraphics[width=0.8\linewidth]{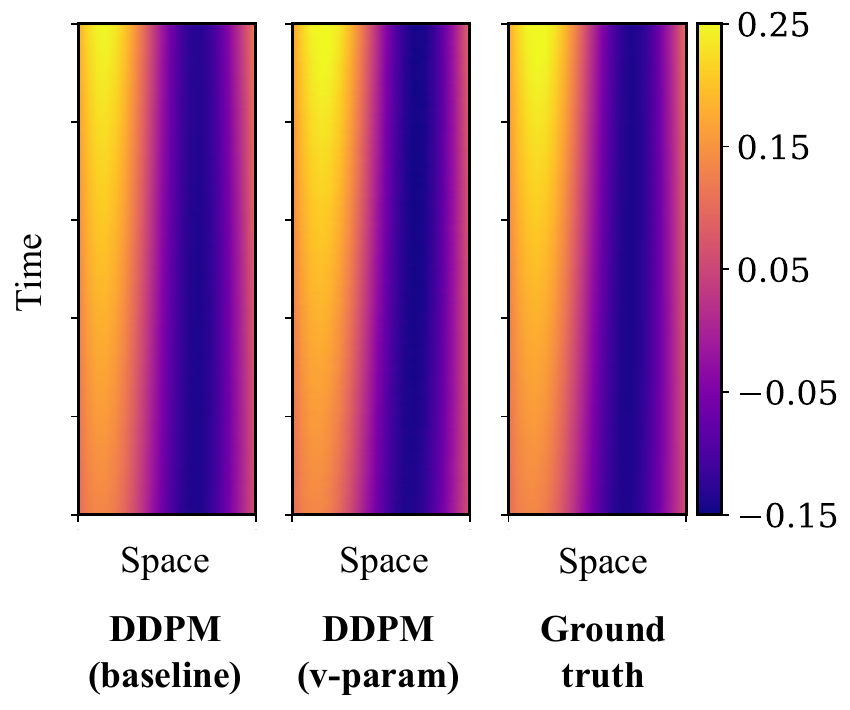}
        \caption{}
        \label{fig:RD_comparisons_u}
    \end{subfigure}
    \hfill
    \begin{subfigure}[b]{0.49\textwidth} \centering
        \includegraphics[width=0.8\linewidth]{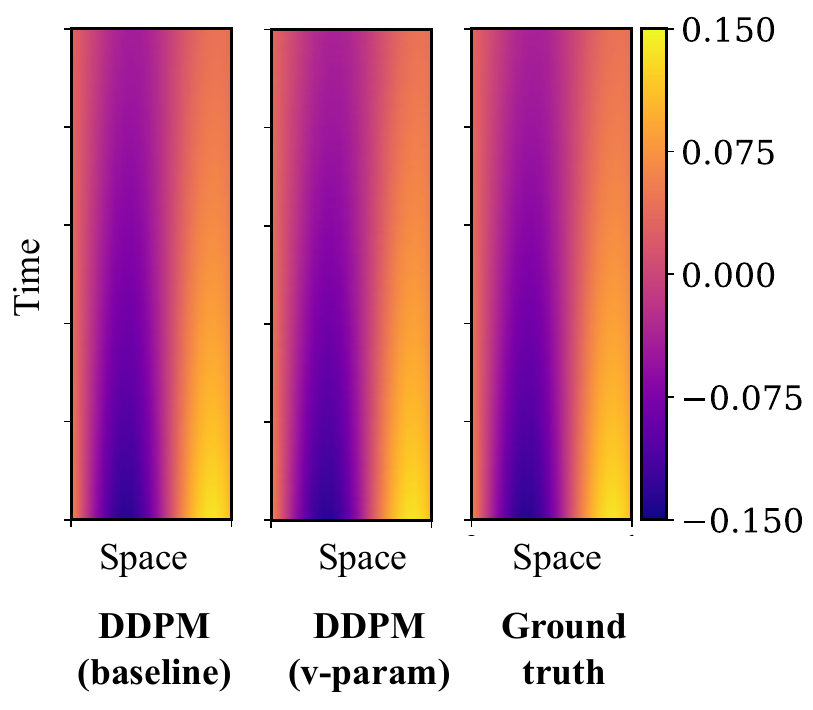}
        \caption{}
        \label{fig:RD_comparisons_v}
    \end{subfigure}
    \caption{Comparison of the compositional DDPM strategies (epsilon baseline and v-parameterization trainings) with the ground truth for predicting the coupled fields in the reaction-diffusion equation: (a) u-field and (b) v-field. Note that the two DDPM models for each test case were only trained with the de-coupled PDE data and were composed together using a PoE strategy to recovery the joint or coupled fields.}
    \label{fig:RD_comparisons}
\end{figure*}

\begin{figure*}[htb!]
    \centering
    \begin{subfigure}[b]{1\textwidth} \centering
        \includegraphics[width=0.8\linewidth]{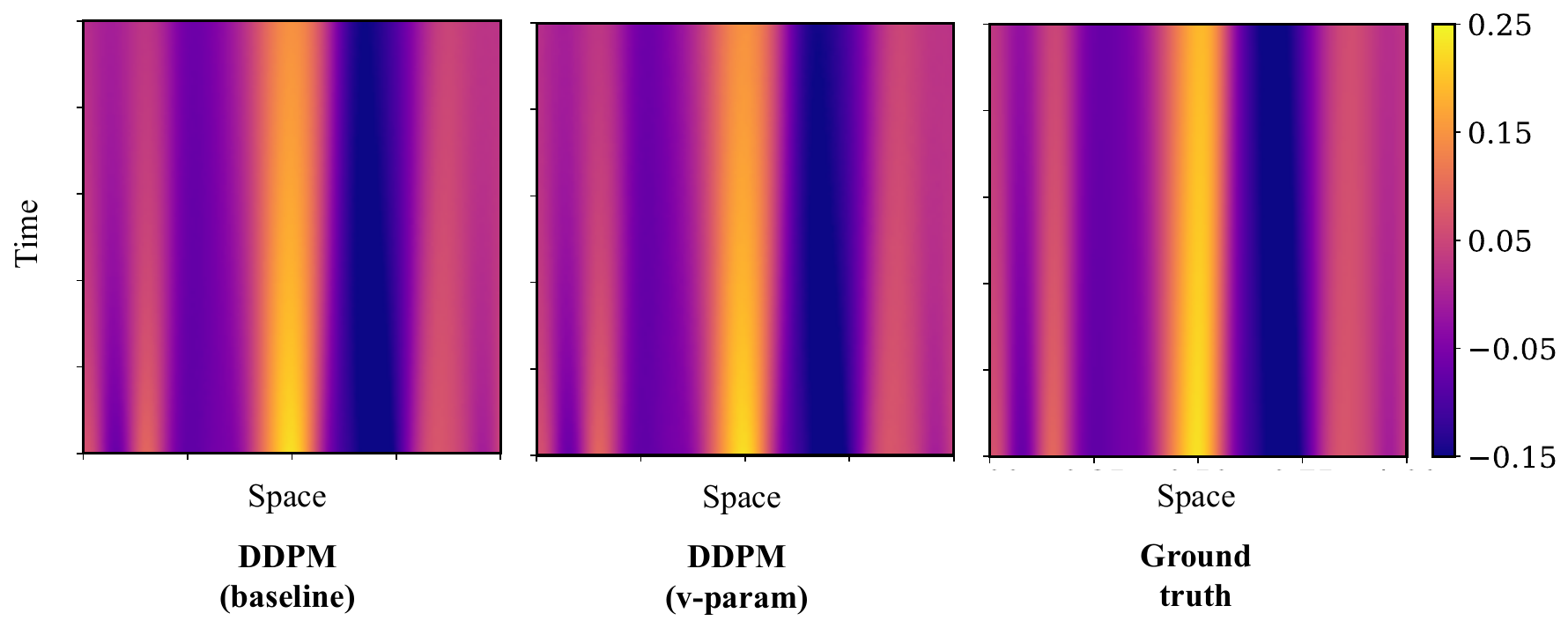}
        \caption{}
        \label{fig:Burger_comparisons_c}
    \end{subfigure}
    \hfill
    \begin{subfigure}[b]{1\textwidth} \centering
        \includegraphics[width=0.8\linewidth]{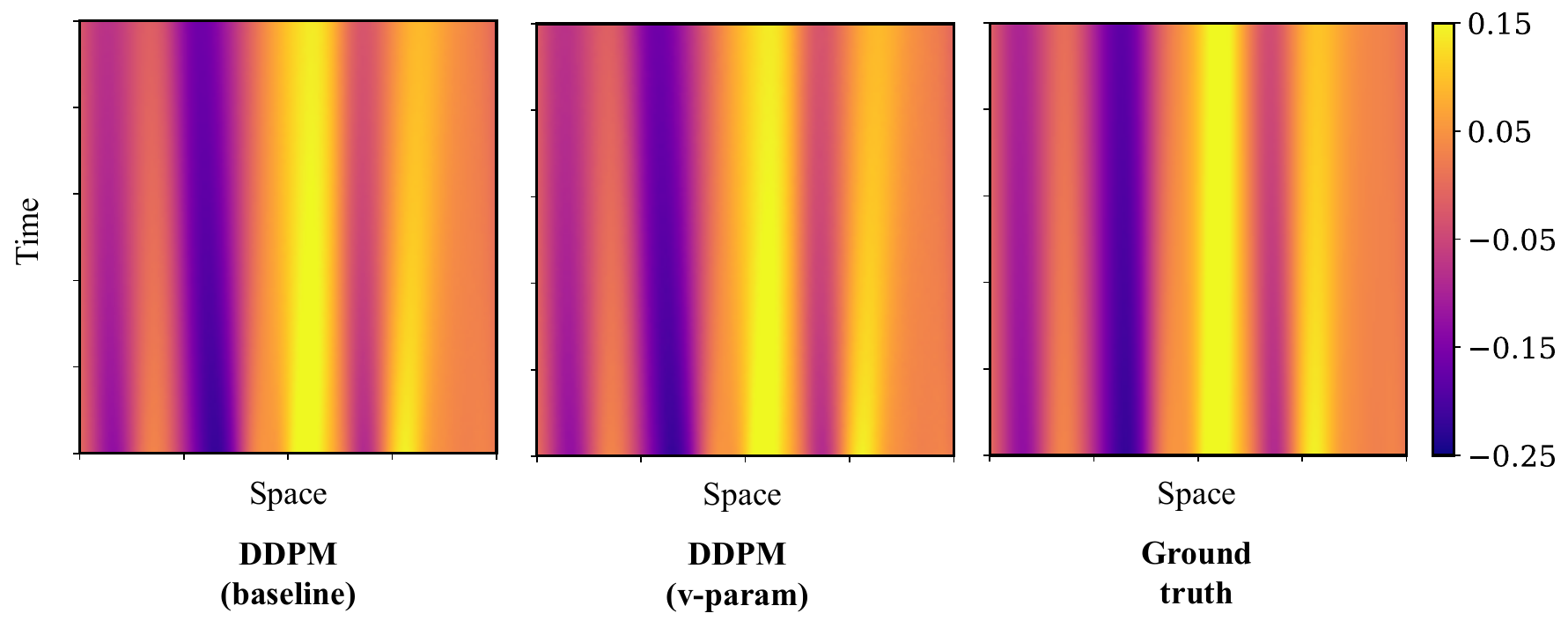}
        \caption{}
        \label{fig:Burger_comparisons_v}
    \end{subfigure}
    \caption{Comparison of the compositional DDPM strategies (epsilon baseline and v-parameterization trainings) with the ground truth for predicting the coupled fields in the modified Burgers equation: (a) c-field and (b) v-field. Note that the two DDPM models for each test case were only trained with the de-coupled PDE data and were composed together using a PoE strategy to recovery the joint or coupled fields.}
    \label{fig:Burger_comparisons}
\end{figure*}

Figures \ref{fig:RD_comparisons_u} and \ref{fig:RD_comparisons_v} show the two coupled fields in the reaction-diffusion equation. The DDPM epsilon strategy (i.e., baseline), DDPM v-parameterization strategy, and the ground truth. These fields are generated for a testing set composed of randomly simulating the initially conditions using a GRF. Visually, the DDPM predicted fields closely match the ground truth for both the fields and under both the parameterizations. Figures \ref{fig:Burger_comparisons_c} and \ref{fig:Burger_comparisons_v} show the two coupled fields in the modified Burgers equation. These fields are generated for a testing set composed of randomly simulating the initially conditions using a GRF. Again, visually, the DDPM predicted fields closely match the ground truth for both the fields and under both the parameterizations. It is again emphasized that the two DDPM models for each test case were only trained with the de-coupled PDE data and were composed together using a PoE strategy to recovery the joint or coupled fields.

Table \ref{tab:test-errors} presents the error metrics on a test dataset for the two compositional diffusion model training strategies. Note that these conditional diffusion models are only trained on the de-coupled data but are used to predict the coupled fields. Also shown for comparison are the test errors corresponding to a FNO trained on the coupled data. It seen first seen that, despite training on the decoupled data, the baseline compositional DDPM performs well. However, its predictive errors are still large. The v-parameterization training strategy is seen to reduce the error considerably for the reaction-diffusion test case. The FNO has the least test error; however, it is directly trained on the coupled data. So, it is expected to perform the best out of the three. 

\begin{table*}[htbp!]
\centering
\caption{Error metrics on a test dataset for the two compositional diffusion model training strategies. Note that these conditional diffusion models are only trained on the de-coupled data but are used to predict the coupled fields. Also shown for comparison are the test errors corresponding to a FNO trained on the coupled data.}
\label{tab:test-errors}
\begin{tabular}{lcc|cc}
\hline
\multicolumn{1}{c}{} & \multicolumn{2}{c}{Reaction--Diffusion (RD)} & \multicolumn{2}{c}{Modified Burgers} \\
\cline{2-5}
Method & MAE $\downarrow$ & RMSE $\downarrow$ & MAE $\downarrow$ & RMSE $\downarrow$ \\
\hline
Fields & (u, v) & (u, v) $\downarrow$ & (c, v) & (c, v) \\
\hline
Compositional DDPM (baseline) & 1.03E-2, 2.5E-3 & 1.49E-2, 3.1E-3 & 9.4E-3, 1.12E-2  & 1.26E-2, 1.54E-2 \\
Compositional DDPM (v-param)  & 4.2E-3, 1.2E-3 & 5.4E-3, 1.5E-3 & 9.8E-3, 1.1E-2 & 1.3E-2, 1.52E-2 \\
FNO trained on coupled data                                & 4E-4, 3E-4 & 7E-4, 4E-4 & 1.1E-3, 1.2E-3 & 1.6E-3, 1.7E-3 \\
\hline
\end{tabular}
\end{table*}

\section{Conclusions and Future Work}

In this paper, compositional diffusion was investigated to compose coupled PDE solutions at inference when the diffusion models are only trained on the uncoupled solutions. Specifically, we were interested in whether the compositional strategy would work when considering long time horizons involving a hundreds of time steps. In addition, we also compared a baseline diffusion model with the v-parameterization strategy. We also introduce a symmetric compositional scheme for the coupled fields based on the Euler scheme. We considered two test cases involving the reaction-diffusion equation and the Burger's equation. We trained a reference FNO model on the coupled data for comparison. Despite seeing only decoupled training data, the compositional diffusion models recover coupled trajectories with low error. v-parameterization can improve accuracy over a baseline diffusion model, while the neural operator surrogate remains strongest given that it is trained on the coupled data. These results show that compositional diffusion is a viable strategy towards efficient, long-horizon modeling of coupled PDEs. Future work will focus on investigating the compositional strategy under tight coupling of the PDEs. It will also expand the compositional diffusion to 3D spaces with long time horizons.

\section{Acknowledgments}
This work is supported through the INL Laboratory Directed Research \& Development (LDRD) Program under DOE Idaho Operations Office Contract DE-AC07-05ID14517.

This research made use of Idaho National Laboratory's High Performance Computing systems located at the Collaborative Computing Center and supported by the Office of Nuclear Energy of the U.S. Department of Energy and the Nuclear Science User Facilities under Contract No. DE-AC07-05ID14517.

\bibliography{aaai2026}

\end{document}